\documentclass[lettersize,journal]{IEEEtran}
\usepackage{amsmath,amsfonts}
\usepackage{algorithmic}
\usepackage{algorithm}
\usepackage{array}
\usepackage[caption=false,font=normalsize,labelfont=sf,textfont=sf]{subfig}
\usepackage{textcomp}
\usepackage{stfloats}
\usepackage{url}
\usepackage{verbatim}
\usepackage{graphicx}
\usepackage{cite}
\usepackage{pifont}
\usepackage{dsfont}
\usepackage{hyperref}

\usepackage{cleveref}
\usepackage{color,soul}
\usepackage[export]{adjustbox}
\usepackage{multirow}
\usepackage[table,xcdraw]{xcolor}

\begin{document}

\title{Continual Self-Supervised Learning with Masked Autoencoders in Remote Sensing}

\author{Lars~Möllenbrok~\IEEEmembership{Student~Member,~IEEE}, Behnood Rasti~\IEEEmembership{Senior~Member,~IEEE}, Begüm~Demir,~\IEEEmembership{Senior~Member,~IEEE}% <-this % stops a space
\thanks{L. Möllenbrok, B. Rasti and B. Demir are with the Faculty of Electrical Engineering and Computer Science, Technische Universit{\"a}t Berlin, 10623 Berlin,
Germany, also with the BIFOLD - Berlin Institute for the Foundations of Learning and Data, 10623 Berlin, Germany. Email: \mbox{lars.moellenbrok@tu-berlin.de}, \mbox{behnood.rasti@tu-berlin.de}, \mbox{demir@tu-berlin.de}.}% <-this % stops a space
}

% The paper headers
\markboth{Journal of \LaTeX\ Class Files,~Vol.~13, No.~9, September~2014}%
{Shell \MakeLowercase{\textit{et al.}}: Bare Demo of IEEEtran.cls for Journals}

\maketitle

\begin{abstract}
 The development of continual learning (CL) methods, which aim to learn new tasks in a sequential manner from the training data acquired continuously, has gained great attention in remote sensing (RS).
 The existing CL methods in RS, while learning new tasks, enhance robustness towards catastrophic forgetting. This is achieved by using a large number of labeled training samples, which is costly and not always feasible to gather in RS. To address this problem, we propose a novel continual self-supervised learning method in the context of masked autoencoders %which have become a popular backbone for self-supervised learning in RS due to their high scalability. 
 (denoted as CoSMAE). The proposed CoSMAE consists of two components: i) data mixup; and ii) model mixup knowledge distillation. Data mixup is associated with retaining information on previous data distributions by interpolating images from the current task with those from the previous tasks. Model mixup knowledge distillation is associated with distilling knowledge from past models and the current model simultaneously by interpolating their model weights to form a teacher for the knowledge distillation. The two components complement each other to regularize the MAE at the data and model levels to facilitate better generalization across tasks and reduce the risk of catastrophic forgetting. %In the experiments, for the continual learning setup we used a challenging scenario of sequential training on four benchmark datasets with different characteristics (which are US3D, UAVid, Potsdam, and TreeSatAI datasets). The proposed method is evaluated on UCMerced data set in terms of mean average precision for multi-label classification. 
Experimental results show that CoSMAE achieves significant improvements of up to $4.94\%$ over state-of-the-art CL methods applied to MAE. Our code is publicly available at: \href{https://git.tu-berlin.de/rsim/CoSMAE}{https://git.tu-berlin.de/rsim/CoSMAE}.
\end{abstract}

\IEEEpeerreviewmaketitle

\section{Introduction}
% The very first letter is a 2 line initial drop letter followed
% by the rest of the first word in caps.
% 
% form to use if the first word consists of a single letter:
% \IEEEPARstart{A}{demo} file is ....
% 
% form to use if you need the single drop letter followed by
% normal text (unknown if ever used by IEEE):
% \IEEEPARstart{A}{}demo file is ....
% 
% Some journals put the first two words in caps:
% \IEEEPARstart{T}{his demo} file is ....
% 
% Here we have the typical use of a "T" for an initial drop letter
% and "HIS" in caps to complete the first word.
\IEEEPARstart  
{T}{he} repeat-pass operations of satellites and the growing number of earth observation missions result in the continuous collection of new remote sensing (RS) images and the frequent release of new training sets. %\hl{These data are a great opportunity for the training of 
Deep neural networks (DNNs) have demonstrated remarkable performance in learning from RS images \mbox{\cite{10669817}}. However, retraining DNNs from scratch every time new data become available poses prohibitive computational challenges, such as high resource consumption and increased training time. On the other hand, current models struggle to learn from data sequentially. Updating a model using only new data can lead to a loss of knowledge from previous (i.e., old) data. This phenomenon is known as catastrophic forgetting \mbox{\cite{9349197}}, which occurs because the data can be non-independently and identically distributed during sequential training. In RS, these distribution shifts can arise due to different acquisition conditions, variations in the spectral response, or changes in land cover. To overcome these problems, continual learning (CL), which aims at learning from data that is made available sequentially (i.e., in tasks), can be used.

 Several CL methods have been developed in RS to address the problem of catastrophic forgetting. For instance, in \mbox{\cite{9444286}}, a method that combines a classifier with a variational autoencoder is presented. The autoencoder learns to generate features from old tasks that are used in the training of the classifier to retain information about the old data distributions. In \mbox{\cite{zhuang2024class}}, a frozen backbone is used together with a linear classifier based on analytic learning. The analytic learning-based classifier enables recursive learning of new data while memorizing old data. 
The method proposed in \mbox{\cite{9513278}} dynamically extends a task-common feature extractor. For each new task, a task-specific classifier is added, together with a lightweight feature transfer module that learns to align the features back to the previous distribution. 
The above-mentioned methods are designed for supervised classification problems and thus require the availability of labeled training images for each CL task\cite{zhuang2024class,9444286,9513278}, which is time-consuming and thus
costly to gather. Accordingly, they are not practical in real
CL applications. An effective approach to address this problem is to exploit self-supervised learning (SSL) that aims to learn general image characteristics without requiring the availability of labeled training data.

According to our knowledge, there is only one continual SSL (CSSL) method in RS \cite{10135093} that combines a contrastive learning method with a regularization based on Elastic Weight Consolidation (EWC) \cite{kirkpatrick2017overcoming}. EWC estimates the importance of model weights after learning a task and discourages changes of important weights when learning a new task. However, as regularization terms accumulate, the method becomes intractable and is not scalable to a large number of tasks. In addition, EWC only relies on the old model weights to retain knowledge of old tasks. However, these old weights might not always optimally capture the knowledge of the old tasks, resulting in sub-optimal performance. 

The development of CSSL methods is much more extended in the computer vision community. As an example, in \mbox{\cite{madaan2022representational}}, 
SSL is combined with an unsupervised data mixup strategy that interpolates images from old tasks and new tasks to alleviate catastrophic forgetting. Since this method relies only on regularization at the data level, performance gain is limited. In \mbox{\cite{gomez2022continually}}, a temporal projector maps representations learned through SSL back to their previous state. The projected representations are then regularized using knowledge distillation from the old model. Although this helps to retain knowledge from the previous tasks, the old model's weights are suboptimal for processing new data, resulting in decreased performance. 

To address the above-mentioned issues, in this letter, we propose a novel CSSL method in the context of masked autoencoders (MAEs) \mbox{\cite{10798628}} which we denote as CoSMAE. The main contributions of CoSMAE lie in its two key components: 1) data mixup introduced for MAE in RS; and 2) model mixup knowledge distillation, which includes a novel strategy that interpolates the weights of the old encoder with those of the current encoder for knowledge distillation.

\section{Proposed CoSMAE Method}
Let $\mathcal{T}_{1:T} = (\mathcal{T}_1,...,\mathcal{T}_T)$ be a sequence of $T$ tasks, where each task $\mathcal{T}_t$ is associated with a set $\mathcal{X}_t=\{X_1^t,...,X_{n_t}^t\}$ of $n_t$ unlabeled RS images. The first task $\mathcal{T}_1$ is associated with an RS image archive $\mathcal{X}_1$, whereas the subsequent tasks $\mathcal{T}_{2:T}$ can be associated with newly acquired RS data or with newly released RS training sets. 
%The data of each task is unique, i.e. $\mathcal{X}_i \cap \mathcal{X}_j = \emptyset$ for $i \neq j$.
We propose a novel CSSL method (denoted as CoSMAE) defined in the context of MAEs, which have recently achieved  significant success in RS image classification.

The proposed CoSMAE aims regularize the MAE during sequential training on $\mathcal{T}_{1:T}$ and to improve generalization across tasks while reducing the risk of catastrophic forgetting.

\begin{figure*}[h]
\centering
\includegraphics[width=17.5cm]{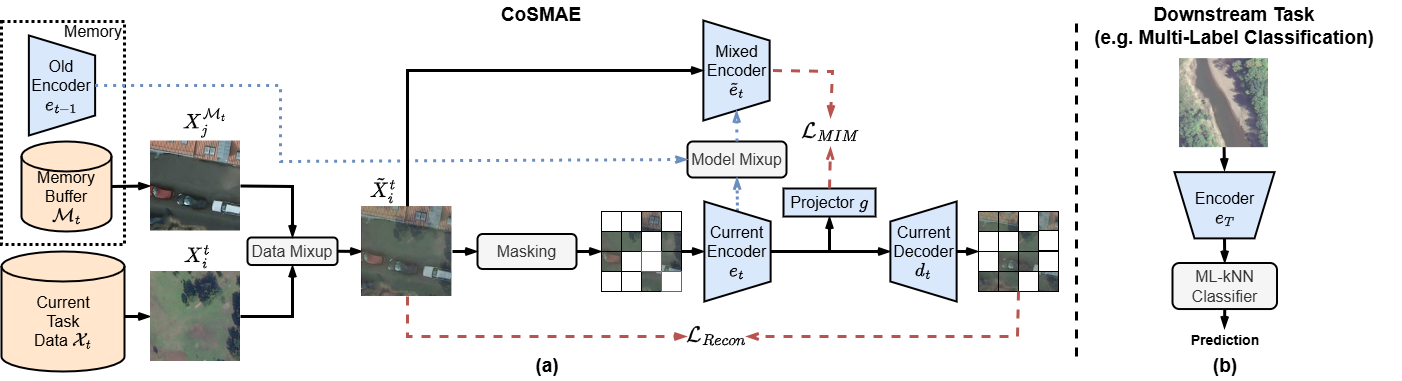}
\caption{(a) An overview of the proposed CoSMAE. It employs data mixup and model mixup knowledge distillation to regularize the model and retain knowledge from previous tasks. Black arrows show the flow of data, dotted blue arrows indicate the flow of model weights, and dashed red arrows indicate the loss computation. (b) Illustration of our downstream task evaluation.}
\label{methodology_overview}
\end{figure*}

To this end, it consists of two components: 1) data mixup; and 2) model mixup knowledge distillation. The former operates on the data level by interpolating new images with old images from a memory buffer, while the latter operates on the model level by interpolating model weights of the current encoder and the weights of the old encoder for knowledge distillation. An overview of CoSMAE is provided in Fig. \ref{methodology_overview}. As the definition of CoSMAE strictly depends on the MAE, in the following subsections we initially present the basics of MAE in the context of CL and then the two key components of CoSMAE are described in detail. 

\subsection{Basics on MAE in the Context of CL}
For an MAE model $f = (e,d)$ that consists of an encoder $e$ and a decoder $d$, let $f_t= (e_t,d_t)$ denote the model trained on tasks $\mathcal{T}_{1:t}$. For  learning a task $\mathcal{T}_{t}$, we consider masked image modeling with patch-wise masking and a Mean Squared Error-based %MSE-based 
reconstruction loss defined as follows:
\begin{equation}
\mathcal{L}_{Recon} = \frac{1}{N_r} \sum_{k=1}^{N_r} \| X^t_{i_k} - \hat{X}^t_{i_k}\|^2,
\label{recon_loss}
\end{equation}
where $X^t_{i_k}$ denotes the $k$-th original patch from $X^t_i$ that is masked and $\hat{X}^t_{i_k}$ its reconstruction learned by the model. The number $N_r$ of masked patches depends on the masking ratio $r$. For learning the first task, (\ref{recon_loss}) is the only objective.
For the learning of subsequent tasks (i.e. $t \geq 2$), we assume to have access to a small memory that consists of a memory buffer and the stored weights of the previous encoder $e_{t-1}$. The memory buffer $\mathcal{M}_t = \{X_1^{\mathcal{M}_t},...,X_{M}^{\mathcal{M}_t}\} $ of size $M$ stores images from previous tasks $\mathcal{T}_{1:t-1}$. We assume that the images from previous tasks are not accessible in line with the CL literature \cite{buzzega2020dark}. Once the MAE is successfully trained, it can be applied to different downstream tasks by transfer learning.
\subsection{Data Mixup}
This component aims to regularize the model at the data level to reduce catastrophic forgetting and to facilitate a better generalization of the learned features. Inspired by \cite{madaan2022representational}, for the first time in RS, we exploit the unsupervised mixup to linearly interpolate images from the current task $\mathcal{T}_{t}$ with images from the memory buffer $\mathcal{M}_t$. In detail, for learning the task $\mathcal{T}_{t}$, instead of training $f_t$ on $\mathcal{X}_{t}$ directly, we train it using images  
 \begin{equation}
      \tilde{X}^t_{i} = \lambda_1 X_i^t + (1-\lambda_1) X^{\mathcal{M}_t}_j,
 \end{equation}
where $\lambda_1 \sim Beta(\alpha_1,\alpha_1)$ for $\alpha_1 \in (0,\infty)$ and $X_j^{\mathcal{M}_t} \in \mathcal{M}_t$ that is randomly selected. It is worth noting that we uniformly select the images from a task to be stored in the memory buffer $\mathcal{M}_t$. The images in the memory buffer are replaced so that the memory buffer contains an equal number of images for each task. 

\subsection{Model Mixup Knowledge Distillation}
This component aims to regularize the current encoder $e_t$ to reduce the risk of catastrophic forgetting and to facilitate better generalization of the learned features. To this end, we introduce a novel knowledge distillation strategy that distills the knowledge learned in previous tasks $\mathcal{T}_{1:t-1}$ and the current task $\mathcal{T}_t$ simultaneously. This is achieved by interpolating the weights $\theta_{t-1}$ of the previous encoder $e_{t-1}$ and the weights $\theta_t$ of the current encoder $e_t$ to obtain the weights $\tilde{\theta}_{t}$ of a mixed encoder $\tilde{e}_t$ serving as a teacher for the distillation of knowledge. In detail, the weights $\tilde{\theta}_{t}$ are obtained by:
\begin{equation}
      \tilde{\theta}_{t} = \lambda_2 \theta_t + (1-\lambda_2) \theta_{t-1},
 \end{equation}
where $\lambda_2 \sim Beta(\alpha_2,\alpha_2)$ for $\alpha_2 \in (0,\infty)$. 
The mixed encoder $\tilde{e}_t$ incorporates both old knowledge and knowledge of the current task. It is noted that the mixed encoder $\tilde{e}_t$ is not trained and does not require any gradient computation.

Let $c^t_i$ denote the output features of the current encoder $e_t$ for the image $\tilde{X}_i^t$. Instead of directly regularizing $c^t_i$, we introduce an additional learnable projector $g$, before applying the regularization (see Fig. \ref{methodology_overview}). It ensures that the current encoder $e_t$ learns the knowledge of the mixed encoder $\tilde{e}_t$ without forcing the output features of $e_{t}$ to be close to those of $\tilde{e}_t$. Forcing the features to be close would reduce the model's ability to adapt to the new data. 
Then, for the knowledge distillation between projected features $g(c_i^t)$ of the current encoder and the output features %$\tilde{e}_t(\tilde{X}^t_i) = \tilde{c}_i^t$ 
$\tilde{c}_i^t$ of the mixed encoder, the mutual information maximization (MIM) loss is used. The MIM loss is based on the normalized temperature-scaled cross-entropy \cite{chen2020simple}. For a batch $\mathcal{B}$ of images it is defined as follows:
\begin{equation}
\begin{split}
    \mathcal{L}_{MIM}(\mathcal{B}) = \frac{1}{2 |\mathcal{B}|} \sum_{\tilde{X}_i^t \in \mathcal{B}} - \log(\frac{e^{S(g(c_i^t),\tilde{c}_i^t)/\tau}}{ \sum_q \mathds{1}_{\{i \neq q\}} e^{S(g(c_i^t),\tilde{c}_q^t)/\tau}}) \\
    - \log(\frac{e^{S(\tilde{c}_i^t,g(c_i^t))/\tau}}{ \sum_q \mathds{1}_{\{i \neq q\}} e^{S(\tilde{c}_i^t, g(c_q^t))/\tau}}),
 \end{split}
\end{equation}
where $S$ denotes the cosine similarity, $\tau$ is the temperature parameter, and $\mathds{1}$ is the indicator function. The loss for training the model is given by:
\begin{equation}
    \mathcal{L} = \mathcal{L}_{Recon} + \beta \mathcal{L}_{MIM},
\end{equation}
where the hyper-parameter $\beta$ determines the contribution of the MIM loss to the total loss. It should be noted that for the mixed encoder, $\tilde{e}_t$ the %full (i.e. unmasked) 
unmasked images are used as input. 

\section{Experimental Results}

\subsection{Dataset Description}
In the experiments, we simulated the sequential availability of new training sets, each of which corresponds to a CL task. For CL, we selected the following datasets: 1) US3D \cite{bosch2019semantic}; 2) UAVid \cite{lyu2020uavid}; 3) Potsdam \cite{rottensteiner2012isprs}; and 4) TreeSatAI \cite{ahlswede2022treesatai}. As a downstream task, we considered multi-label classification (MLC) applied to the UCMerced dataset \cite{yang2012geographic}. 
 %All the considered datasets are explained in the following. 
%\subsubsection{US3D \textnormal{\cite{bosch2019semantic}}}
The US3D dataset contains images acquired by the WorldView-3 satellite over three different US cities. We selected a subset of images from Jacksonville, Nebraska. Each image has eight bands with a ground sample distance of 30cm. We divided the images into non-overlapping image patches of size 256 $\times$ 256 pixels. Then, we randomly selected a subset of 21081 patches from the original training set for training while choosing 3072 patches from the original validation set for validation. 
The UAVid dataset contains RGB images gathered from different video sequences captured from an oblique point of view. Each image has a size of 3840 × 2160 pixels. We selected the images from the original training and validation sets and divided them into non-overlapping image patches of size 256 × 256 pixels, resulting in 24560 patches for training and 8640 patches for validation. 
%\subsubsection{Potsdam \textnormal{\cite{rottensteiner2012isprs}}}
 The Potsdam dataset consists of 38 aerial true orthophotos of size 6000 $\times$ 6000 pixels with four bands. The images have a spatial resolution of 5cm. For the experiments, we divided the images into non-overlapping image patches of size 256 $\times$ 256 pixels. We randomly selected 15870 patches for training and 1058 patches for validation. %3174/
%\subsubsection{TreeSatAI \textnormal{\cite{ahlswede2022treesatai}}}
The TreeSatAI dataset contains 50380 aerial images that are of size 300 $\times$ 300 pixels and have four bands. The images have a spatial resolution of 20 cm. We randomly selected a subset of 15000 images for training and 1000 for validation.
%\subsubsection{UCMerced \textnormal{\cite{yang2012geographic}}}
The UCMerced dataset consists of 2100 RGB images of size 256 $\times$ 256 pixels with a spatial resolution of 30cm. %We used multi-labels presented in \cite{chaudhuri2017multilabel}. 
We randomly selected 1000 images for training, 300 images for validation, and 800 images for testing. For both training and evaluation, we used the multi-labels of images that are described in \mbox{\cite{chaudhuri2017multilabel}}.

\subsection{Design of Experiments}

\begin{figure*}[htp]
  \centering
  \begin{minipage}[b]{0.329\textwidth}
  %\begin{subfigure}[]
\subfloat[]{
\includegraphics[width=\textwidth]{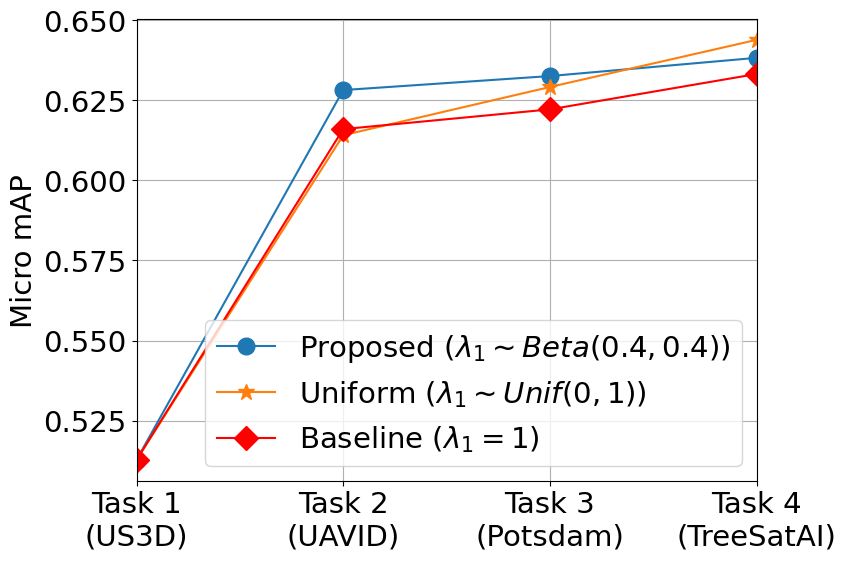}
\label{datamixup}}
\end{minipage}
    %}
    %\caption{test}
  %\end{subfigure}
\hfill
\begin{minipage}[b]{0.329\textwidth}
  %\begin{subfigure}[]
   % \centering
    \subfloat[]{
    \includegraphics[width=\textwidth]{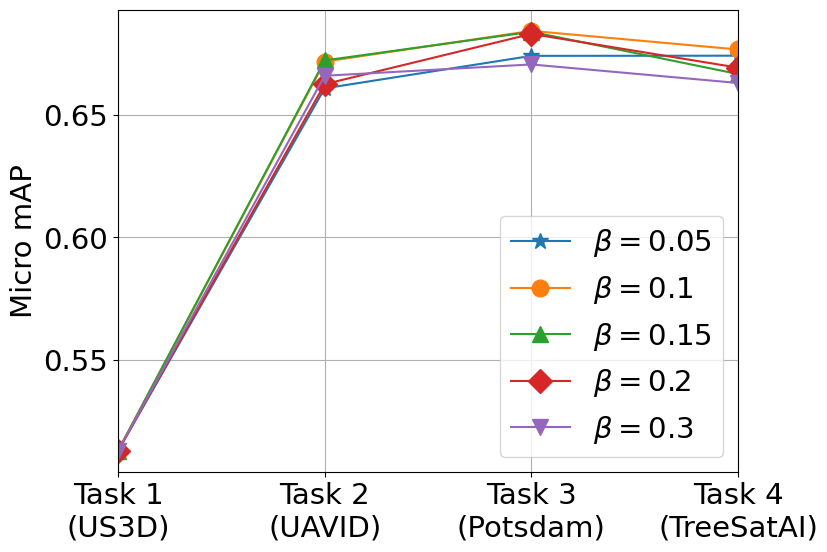}
    \label{beta_param}}
\end{minipage}
    %}
    %\caption{test}
  %\end{subfigure}
  \hfill
  %\begin{subfigure}[]
    %\centering
\begin{minipage}[b]{0.33\textwidth}
    \subfloat[]{
    \includegraphics[width=\textwidth]{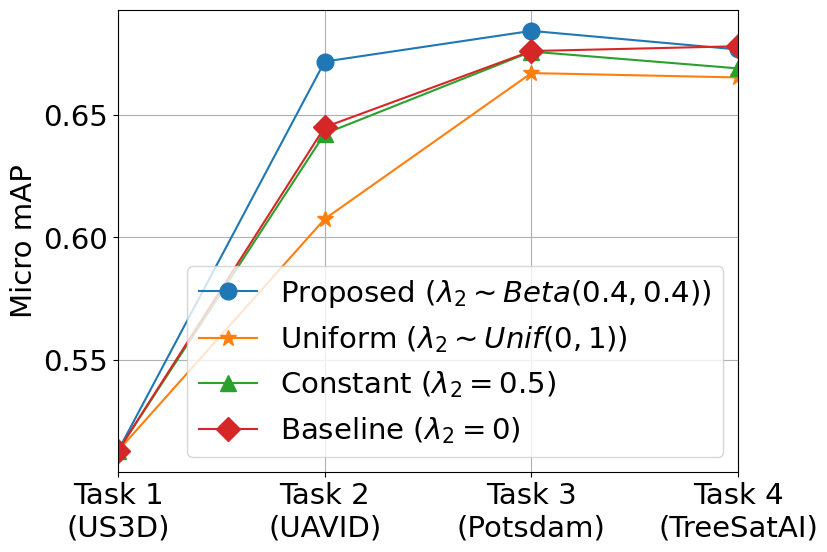}
    \label{modelmixup}}
\end{minipage}
    %}
    %\caption{test}
  %\end{subfigure}
  %\end{minipage}
  %\hfill
  %\begin{minipage}[b]{0.329\textwidth}
  %  \centering
  %  \subfloat[]{\includegraphics[width=\textwidth]{figures/kd_beta_micro_big_marker_cut_smaller_2.png}
  %  \label{beta_param}}
  %\end{minipage}
  %\hfill
  %\begin{minipage}[b]{0.33\textwidth}
  %  \centering
  %  \subfloat[]{\includegraphics[width=\textwidth]{figures/model_mixup_KD_micro_big_marker_cut_smaller_2.png}
  %  \label{modelmixup}}
  %\end{minipage}
    \caption{Micro mAP results obtained on UCMerced associated to each task of sequential training for the assessment of: (a) the proposed data mixup, uniform data mixup, and the baseline; (b) different values of $\beta$; and (c) the proposed model mixup knowledge distillation, uniform model mixup, constant model mixup, and baseline.}
  %\caption{MLC results in terms of micro mAP on UCMerced after sequential training up to a specific task. Analyzing: (a) the performance of the proposed data mixup, uniform data mixup, and no data mixup (Baseline) for MAE; (b) the effect of different values of $\beta$; and (c) the performance of the proposed model mixup knowledge distillation, uniform model mixup, constant model mixup, and using the old encoder without model mixup (Baseline) for MAE.}
\end{figure*}

As the backbone of the MAE architecture, we used ViT models for both the encoder and the decoder. The encoder consists of 12 blocks, 12 heads, a patch size of 16, and a hidden dimension of 768. The decoder uses 8 blocks, 16 attention heads, and an embedding dimension of 512. The masking ratio $r$ was set to 0.75. For the projector $g$, we used a two-layer MLP with a hidden dimension of 128 and ReLU activation. We used random horizontal flipping and random resized cropping to 224 $\times$ 224 pixels with a minimum crop ratio of 0.8. For each task, we trained for 300 epochs with a batch size of 128. We used AdamW optimizer with a learning rate of $1e^{-3}$, linear warmup for 10 epochs, and cosine annealing.
The size $M$ of the memory buffer $\mathcal{M}_t$ was set to 1000. The temperature $\tau$ in the MIM loss was set to 0.5. For evaluation, the learned encoder is used as a feature extractor in the evaluation data set, and a multi-label $k$-Nearest-Neighbor (ML-$k$NN) classifier with $k=10$ is used to classify the extracted representations. The results are provided in terms of micro mean average precision (mAP). Note that we have observed similar behavior in results in terms of macro mAP (not reported in the paper due to space constraints). In our experiments, we used the RGB bands of each dataset. All experimental results are based on the average micro mAP obtained in three trials using three randomly initialized models.  All experiments were carried out on a single NVIDIA A100 GPU. %All the results were obtained by averaging three trials. 

\subsection{Ablation and Sensitivity Study}
In this subsection, first we present an ablation study to analyze the impact of the two components of the proposed method. To this end, we present the results %This is done by looking at different configurations 
obtained by excluding individual components of the proposed method. When data mixup is excluded, the model is trained only on the data of the current task without the interpolation of images. When model mixup knowledge distillation is excluded, we only use the reconstruction loss for training without knowledge distillation with the mixed encoder. Table \mbox{\ref{ablation_table}} shows the micro mAP obtained for the classification of UCMerced after completing the training up to a specific task within the sequence of CL tasks. From the table, it is evident that the highest micro mAP is obtained when both data mixup and model mixup knowledge distillation are used. Furthermore, we observe that model mixup knowledge distillation has a greater impact on the performance with respect to the data mixup.

\begin{table}[h]
    \centering
    \caption{Ablation study: Micro mAP (\%) obtained using different components of the proposed CoSMAE on UCMerced associated with each task of sequential training.}
    \begin{tabular}{|>{\centering\arraybackslash}m{0.7cm}|>
{\centering\arraybackslash}m{1.25cm}|>
{\centering\arraybackslash}m{0.8cm}|>{\centering\arraybackslash}m{0.95cm}|>{\centering\arraybackslash}m{1.18cm}|>{\centering\arraybackslash}m{1.35cm}|}
    \hline
        Data Mixup & Model Mixup KD  & Task 1 (US3D)& Task 2 (UAVID) & Task 3 (Potsdam)& Task 4 (TreeSatAI)\\
        \hline 
         \hline 
        \ding{55} & \ding{55} &  \multirow{4}{*}{51.28} & 60.33 & 61.29 & 61.12\\
          \ding{51} & \ding{55} &  & 62.82 & 63.25 & 63.83  \\
          \ding{55} & \ding{51} &   & 65.59 & 67.17 & 66.93 \\
         \ding{51} & \ding{51} &  & \textbf{67.19} & \textbf{68.44} & \textbf{67.68}  \\
        
        \hline
    \end{tabular}

    \label{ablation_table}
\end{table}

We also analyze the effect of data mixup under different sampling strategies for the interpolation parameter $\lambda_1$ without applying model mixup knowledge distillation. Fig. \ref{datamixup} shows the micro mAP after completing the training up to a specific task within the sequence of CL tasks. From the figure, we can see that the proposed data mixup strategy ($\lambda_1 \sim Beta(0.4,0.4)$) results in the highest micro mAP for all tasks. Data mixup based on sampling from a uniform distribution ($\lambda_1 \sim \mathcal{U}_{[0,1]}$) shows lower performance on the second and third tasks, but slightly outperforms the proposed data mixup on the final task. In the proposed data mixup, $\lambda_1$ is %likely to be 
selected close to $0$ or $1$ with high probability. %(see Fig. \ref{beta} for a visualization of the beta density function). 
In this way, data mixup can simultaneously serve as a replay of information from old data and a data augmentation strategy that regularizes the model.  %(see Fig. \ref{data_mixup_examples} for example images). 
This results in retaining knowledge of old data and better generalization of the model. When applying a uniform sampling strategy, $\lambda_1$ takes all values between 0 and 1 with equal probability. Therefore, there are many interpolated images for values of $\lambda_1$ that are not close to 0 or 1. These images follow a data distribution that is very different from the task's data distribution, caused by the overlapping of different shapes and textures. Consequently, the model shows degraded performance. The different behavior on the last task could be attributed to the different properties of the TreeSatAI dataset compared to the other datasets. %In detail, it includes images with tree-species classes. When interpolated with images from the other tasks, the original shapes and textures are often still discernible.
 We also analyze the effect of model mixup knowledge distillation with respect to $\beta$ to determine the contribution of the MIM term compared to the reconstruction loss. From Fig. \ref{beta_param}, one can see that for $\beta=0.1$ the highest mAP is achieved. Finally, we analyze the effect of model mixup knowledge distillation under different sampling strategies for the interpolation parameter $\lambda_2$ (see Fig. \ref{modelmixup}). In this case, for all models we use data mixup, the same projector, and the mutual information maximization loss for knowledge distillation with $\beta=0.1$. From Fig. \ref{modelmixup}, we can observe that the proposed model mixup knowledge distillation ($\lambda_2 \sim Beta(0.4,0.4)$) results in the overall highest micro mAP across the tasks. Using a constant interpolation parameter of $\lambda_2=0.5$ performs similarly to the baseline ($\lambda_2 = 0$) while sampling $\lambda_2$ uniformly ($\lambda_2 \sim \mathcal{U}_{[0,1]}$) degrades the performance. In the proposed model mixup knowledge distillation, $\lambda_2$ is %likely to be
 selected to be close to $0$ or $1$ with high probability. Therefore, the mixed encoder can regularize the current encoder with knowledge from the old and the current tasks. The model is forced to learn features that generalize to both types of regularization, resulting in a more robust model. When applying a uniform sampling strategy for $\lambda_2$ ($\lambda_2 \sim \mathcal{U}_{[0,1]}$), the mixed encoder takes on any weight composition across the linear interpolation between the old encoder and current encoder with equal probability. This causes the model to be overly constrained by the regularization, resulting in poor performance across all tasks. 

\subsection{Comparison with State-of-the-Art CL Methods}
In this subsection, we compare CoSMAE with CL strategies from the state-of-the-art (Experience Replay [ER] \cite{robins1995catastrophic}, EWC \cite{kirkpatrick2017overcoming}, Dark Experience Replay [DER]\cite{buzzega2020dark}, Lifelong Unsupervised Mixup [LUMP]\cite{madaan2022representational}, Projected Functional Regularization [PFR] \mbox{\cite{gomez2022continually}}) applied to MAE and also a baseline (which trains an MAE sequentially based on the tasks without applying a CL strategy). DER was originally proposed for supervised CL problems. We adapted it to be self-supervised by replacing the regularization of the memory image logits with the regularization of the encoded features as proposed in \cite{madaan2022representational}. Table \mbox{\ref{soa_table}} shows the micro mAP %of the ML-$k$NN classifier evaluated on the extracted representations of UCMerced, 
after completing the training up to a specific task within the sequence of CL tasks. Additionally, Table \mbox{\ref{soa_table}} shows the total training times required to complete the entire sequence of tasks. It should be noted that, in the first task, the training is the same for all the methods since no CL is employed at this stage. From the table, we can see that CoSMAE outperforms the other methods by a significant margin. As an example, for the second task, CoSMAE achieves a micro mAP of $66.71\%$, which is $4.94\%$ higher than the %highest among the 
state-of-the-art and $6.38\%$ higher than the baseline. For the third task, the proposed method outperforms %the highest among 
the state-of-the-art by $2.04\%$ and the baseline by $6.75\%$. In the last task, the proposed method outperforms the state-of-the-art by $0.94\%$ and the baseline by $6.14\%$. Most of the state-of-the-art methods show only minor improvements over the baseline. 
From the table, we can also see that 
CoSMAE has a total training time of $13.77$ hours, which is $3.69$ hours more than the baseline and $0.84$ hours more than the second-best performing method PFR. The increased training times of CoSMAE, PFR, and DER are attributed to the knowledge distillation, which requires an additional forward pass during training. We would like to note that the baseline is expected to have the lowest training time, since no additional strategies for retaining old knowledge are employed.

\begin{table}
    \centering
    \caption{Micro mAP (\%) obtained on UCMerced associated with each task of sequential training and the training time (hours) for the proposed method, the state-of-the-art methods, and the baseline.}
    \begin{tabular}{|>{\centering\arraybackslash}m{1.2cm}|>{\centering\arraybackslash}m{0.77cm}|>{\centering\arraybackslash}m{0.9cm}|>{\centering\arraybackslash}m{1.16cm}|>{\centering\arraybackslash}m{1.35cm}| >{\centering\arraybackslash}m{0.9cm}|}
    \hline
     & \multicolumn{4}{c|}{micro mAP} &  Total \\
     \cline{2-5}
      Method \ \ \ \ \ & Task 1 (US3D)& Task 2 (UAVID) & Task 3 (Potsdam)& Task 4 (TreeSatAI) &  training time \\
        \hline 
         \hline 
        Baseline & \multirow{7}{*}{49.11} & 60.33 & 61.29 & 61.12 &  \textbf{10.08} \\
        ER \cite{robins1995catastrophic} &  & 61.18 & 61.72 & 62.57 &  10.41 \\
        EWC \cite{kirkpatrick2017overcoming} &  & 60.93 & 61.68 & 63.06 &  10.9 \\
        DER \cite{buzzega2020dark} &  & 61.75 & 61.69 & 61.26 &  11.92 \\
        LUMP \cite{madaan2022representational} &  & 61.65 & 61.53 & 62.37 &  10.16 \\
         PFR \cite{gomez2022continually} &  & 61.77 & 66.00 & 66.32 &  12.93 \\
         
CoSMAE &  & \textbf{66.71} & \textbf{68.04} & \textbf{67.26} &  13.77 \\
        
        \hline
    \end{tabular}
    \label{soa_table}
\end{table}

\section{Conclusion}
In this letter, we have proposed a continual self-supervised  learning method (denoted CoSMAE) in RS. The proposed method consists of two components: i) data mixup; and ii) model mixup knowledge distillation. The joint consideration of these components reduces the risk of catastrophic forgetting, while increasing the generalization capability of the model to the downstream task. 
From the experiments, we observed that CoSMAE outperforms the state-of-the-art CL methods applied in the context of MAE at the cost of a slight increase in training time. In particular, its novel component (which is model mixup knowledge distillation) contributes to a large increase in performance by forming a strong teacher based on interpolated model weights that allows the model to learn more discriminative features. 
It is worth emphasizing that, although we evaluated the proposed method in MLC problems, using RGB datasets, it is independent of the considered learning task (and can thus be used with any task by using a task-specific head) as well as the data modality. As a future work, we plan to exploit our method for sequential training of multi-modal data acquired by different sensors (e.g., multispectral, hyperspectral) under different learning tasks (e.g., semantic segmentation).

\bibliographystyle{IEEEtran}
\bibliography{references.bib}

% Generated by IEEEtran.bst, version: 1.14 (2015/08/26)
\begin{thebibliography}{10}
\providecommand{\url}[1]{#1}
\csname url@samestyle\endcsname
\providecommand{\newblock}{\relax}
\providecommand{\bibinfo}[2]{#2}
\providecommand{\BIBentrySTDinterwordspacing}{\spaceskip=0pt\relax}
\providecommand{\BIBentryALTinterwordstretchfactor}{4}
\providecommand{\BIBentryALTinterwordspacing}{\spaceskip=\fontdimen2\font plus
\BIBentryALTinterwordstretchfactor\fontdimen3\font minus \fontdimen4\font\relax}
\providecommand{\BIBforeignlanguage}[2]{{%
\expandafter\ifx\csname l@#1\endcsname\relax
\typeout{** WARNING: IEEEtran.bst: No hyphenation pattern has been}%
\typeout{** loaded for the language `#1'. Using the pattern for}%
\typeout{** the default language instead.}%
\else
\language=\csname l@#1\endcsname
\fi
#2}}
\providecommand{\BIBdecl}{\relax}
\BIBdecl

\bibitem{10669817}
D.~Tuia, K.~Schindler, B.~Demir, X.~X. Zhu, M.~Kochupillai, S.~Džeroski, J.~N. van Rijn, H.~H. Hoos, F.~Del~Frate, M.~Datcu, V.~Markl, B.~Le~Saux, R.~Schneider, and G.~Camps-Valls, ``Artificial intelligence to advance earth observation: A review of models, recent trends, and pathways forward,'' \emph{IEEE Geoscience and Remote Sensing Magazine}, pp. 2--25, 2024.

\bibitem{9349197}
M.~De~Lange, R.~Aljundi, M.~Masana, S.~Parisot, X.~Jia, A.~Leonardis, G.~Slabaugh, and T.~Tuytelaars, ``A continual learning survey: Defying forgetting in classification tasks,'' \emph{IEEE Transactions on Pattern Analysis and Machine Intelligence}, vol.~44, no.~7, pp. 3366--3385, 2022.

\bibitem{9444286}
N.~Ammour, ``Continual learning using data regeneration for remote sensing scene classification,'' \emph{IEEE Geoscience and Remote Sensing Letters}, vol.~19, pp. 1--5, 2022.

\bibitem{zhuang2024class}
H.~Zhuang, Y.~Yan, R.~He, and Z.~Zeng, ``Class incremental learning with analytic learning for hyperspectral image classification,'' \emph{Journal of the Franklin Institute}, vol. 361, no.~18, p. 107285, 2024.

\bibitem{9513278}
X.~Lu, X.~Sun, W.~Diao, Y.~Feng, P.~Wang, and K.~Fu, ``L{IL}: Lightweight incremental learning approach through feature transfer for remote sensing image scene classification,'' \emph{IEEE Transactions on Geoscience and Remote Sensing}, vol.~60, pp. 1--20, 2022.

\bibitem{10135093}
V.~Marsocci and S.~Scardapane, ``Continual barlow twins: Continual self-supervised learning for remote sensing semantic segmentation,'' \emph{IEEE Journal of Selected Topics in Applied Earth Observations and Remote Sensing}, vol.~16, pp. 5049--5060, 2023.

\bibitem{kirkpatrick2017overcoming}
J.~Kirkpatrick, R.~Pascanu, N.~Rabinowitz, J.~Veness, G.~Desjardins, A.~A. Rusu, K.~Milan, J.~Quan, T.~Ramalho, A.~Grabska-Barwinska \emph{et~al.}, ``Overcoming catastrophic forgetting in neural networks,'' \emph{Proceedings of the National Academy of Sciences}, vol. 114, no.~13, pp. 3521--3526, 2017.

\bibitem{madaan2022representational}
D.~Madaan, J.~Yoon, Y.~Li, Y.~Liu, and S.~J. Hwang, ``Representational continuity for unsupervised continual learning,'' in \emph{International Conference on Learning Representations}, 2022.

\bibitem{gomez2022continually}
A.~Gomez-Villa, B.~Twardowski, L.~Yu, A.~D. Bagdanov, and J.~Van~de Weijer, ``Continually learning self-supervised representations with projected functional regularization,'' in \emph{Proceedings of the IEEE/CVF Conference on Computer Vision and Pattern Recognition}, 2022, pp. 3867--3877.

\bibitem{10798628}
J.~Hackstein, G.~Sumbul, K.~Norman~Clasen, and B.~Demir, ``Exploring masked autoencoders for sensor-agnostic image retrieval in remote sensing,'' \emph{IEEE Transactions on Geoscience and Remote Sensing}, vol.~63, pp. 1--14, 2025.

\bibitem{buzzega2020dark}
P.~Buzzega, M.~Boschini, A.~Porrello, D.~Abati, and S.~Calderara, ``Dark experience for general continual learning: a strong, simple baseline,'' \emph{Advances in neural information processing systems}, vol.~33, pp. 15\,920--15\,930, 2020.

\bibitem{chen2020simple}
T.~Chen, S.~Kornblith, M.~Norouzi, and G.~Hinton, ``A simple framework for contrastive learning of visual representations,'' \emph{International conference on machine learning}, pp. 1597--1607, 2020.

\bibitem{bosch2019semantic}
M.~Bosch, K.~Foster, G.~Christie, S.~Wang, G.~D. Hager, and M.~Brown, ``Semantic stereo for incidental satellite images,'' \emph{IEEE Winter Conference on Applications of Computer Vision}, pp. 1524--1532, 2019.

\bibitem{lyu2020uavid}
Y.~Lyu, G.~Vosselman, G.-S. Xia, A.~Yilmaz, and M.~Y. Yang, ``U{AV}id: A semantic segmentation dataset for {UAV} imagery,'' \emph{ISPRS journal of photogrammetry and remote sensing}, vol. 165, pp. 108--119, 2020.

\bibitem{rottensteiner2012isprs}
F.~Rottensteiner, G.~Sohn, J.~Jung, M.~Gerke, C.~Baillard, S.~Benitez, and U.~Breitkopf, ``The {ISPRS} benchmark on urban object classification and 3{D} building reconstruction,'' \emph{ISPRS Annals of the Photogrammetry, Remote Sensing and Spatial Information Sciences; I-3}, vol.~1, no.~1, pp. 293--298, 2012.

\bibitem{ahlswede2022treesatai}
S.~Ahlswede, C.~Schulz, C.~Gava, P.~Helber, B.~Bischke, M.~F{\"o}rster, F.~Arias, J.~Hees, B.~Demir, and B.~Kleinschmit, ``Tree{S}at{AI} benchmark archive: A multi-sensor, multi-label dataset for tree species classification in remote sensing,'' \emph{Earth System Science Data Discussions}, pp. 1--22, 2022.

\bibitem{yang2012geographic}
Y.~Yang and S.~Newsam, ``Geographic image retrieval using local invariant features,'' \emph{IEEE Transactions on Geoscience and Remote Sensing}, vol.~51, no.~2, pp. 818--832, 2012.

\bibitem{chaudhuri2017multilabel}
B.~Chaudhuri, B.~Demir, S.~Chaudhuri, and L.~Bruzzone, ``Multilabel remote sensing image retrieval using a semisupervised graph-theoretic method,'' \emph{IEEE Transactions on Geoscience and Remote Sensing}, vol.~56, no.~2, pp. 1144--1158, 2017.

\bibitem{robins1995catastrophic}
A.~Robins, ``Catastrophic forgetting, rehearsal and pseudorehearsal,'' \emph{Connection Science}, vol.~7, no.~2, pp. 123--146, 1995.

\end{thebibliography}
\end{document}